%% file: main.tex
\documentclass[10pt,twocolumn,letterpaper]{article}

\usepackage{cvpr}              

\input{preamble}

\usepackage{multirow}
\usepackage[accsupp]{axessibility} 
\definecolor{cvprblue}{rgb}{0.21,0.49,0.74}
\usepackage[pagebackref,breaklinks,colorlinks,citecolor=cvprblue]{hyperref}

\def\paperID{3504} 
\def\confName{CVPR}
\def\confYear{2024}

\newcommand\blfootnote[1]{%
  \begingroup
  \renewcommand\thefootnote{}\footnote{#1}%
  \addtocounter{footnote}{-1}%
  \endgroup
}

\title{LLMs are Good Action Recognizers}

\author{Haoxuan Qu$^{1}$
~~~ Yujun Cai$^{2}$
~~~ Jun Liu$^{1\dag}$ \\
\textsuperscript{1}Singapore University of Technology and Design ~~ \textsuperscript{2}Nanyang Technological University \\
{\tt\small haoxuan\_qu@mymail.sutd.edu.sg, yujun001@e.ntu.edu.sg, jun\_liu@sutd.edu.sg } \\
}

\begin{document}
\maketitle

\blfootnote{\dag~Corresponding author}

\input{sec/0_abstract}    
\input{sec/1_intro}

\input{sec/2_related_work}

\input{sec/3_method}
\input{sec/4_experiments}
\input{sec/5_conclusion}

\noindent\textbf{Acknowledgement.} This project is supported by the Ministry of Education, Singapore, under the AcRF Tier 2 Projects (MOE-T2EP20222-0009 and MOE-T2EP20123-0014).

\clearpage
{
    \small
    \bibliographystyle{ieeenat_fullname}
    \bibliography{main}
}


\end{document}

%% file: preamble.tex
\usepackage{graphicx}
\usepackage{wrapfig}
\usepackage[dvipsnames]{xcolor}


%% file: sec/0_abstract.tex
\begin{abstract}
Skeleton-based action recognition has attracted lots of research attention. 
Recently, to build an accurate skeleton-based action recognizer, a variety of works have been proposed. Among them, some works use large model architectures as backbones of their recognizers to boost the skeleton data representation capability, while some other works pre-train their recognizers on external data to enrich the knowledge. 
In this work, we observe that large language models which have been extensively used in various natural language processing tasks generally hold both large model architectures and rich implicit knowledge. 
Motivated by this, we propose a novel \textbf{LLM-AR} framework, in which we investigate treating the \textbf{L}arge \textbf{L}anguage \textbf{M}odel as an \textbf{A}ction \textbf{R}ecognizer. In our framework, we propose a linguistic projection process to project each input action signal (i.e., each skeleton sequence) into its ``sentence format'' (i.e., an ``action sentence''). 
Moreover, we also incorporate our framework with several designs to further facilitate this linguistic projection process.
Extensive experiments demonstrate the efficacy of our proposed framework.
\end{abstract}

%% file: sec/1_intro.tex
\section{Introduction}
\label{sec:intro}

Human action recognition aims to categorize the actions performed by humans into a pre-defined list of classes. It is relevant to a variety of applications, such as human-computer interaction \cite{liu2017human}, intelligent surveillance \cite{lin2008activity}, and virtual reality \cite{fangbemi2018efficient}. 
In the past few years, skeleton-based action recognition \cite{yan2018spatial,shi2019skeleton,shi2019two,cheng2020decoupling,chen2021channel,cai2023ske2grid,wang2023neural,duan2023skeletr} has received a lot of research attention with the notice that skeleton is a succinct yet informative representation of human behaviors.
Yet, despite the considerable progress, skeleton-based action recognition still remains a challenging task \cite{zhou2023learning}, and to build a more accurate skeleton-based action recognizer, various recent works have been proposed from different perspectives.
Among them, some recent works \cite{wang20233mformer,foo2023unified} proposed to utilize \textit{large model architectures} (such as transformers) as the backbone of their action recognizers to achieve stronger representation capability and capture more subtle differences among different actions. 
On the other hand, some other works \cite{motionbert2022,yang2021unik} proposed to pre-train their action recognizers on external data 
in order for their action recognizers to handle this task with richer knowledge.

\begin{figure}[t]
\centering
\includegraphics[width=\columnwidth]{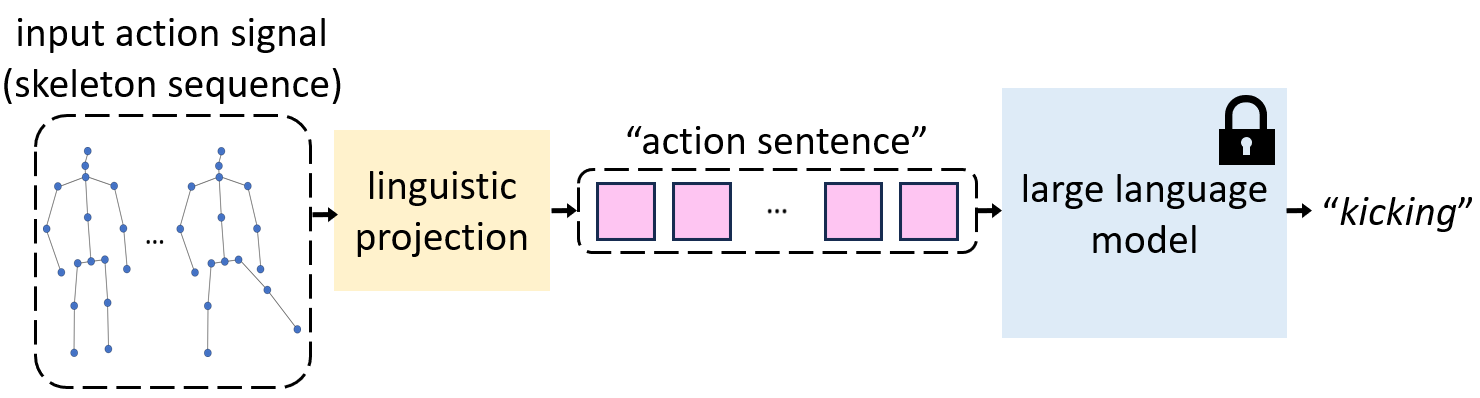}
\vspace{-0.5cm}
\caption{Overview of our proposed LLM-AR framework. In our framework, given an input action signal, we first perform a linguistic projection process to acquire the corresponding ``action sentence''. We then perform action recognition via the large language model with its pre-trained weights untouched to keep its pre-learned rich knowledge.}
\vspace{-0.5cm}
\label{fig:intro}
\end{figure}

Recently, large language models such as GPT \cite{brown2020language} and LLaMA \cite{touvron2023llama} have become quite popular and have been extensively applied in handling various human languages. 
For example, having been pre-trained over various human languages and learned common language-related characteristics, large language models can be very effectively and efficiently adapted to even new human languages unseen during pre-training \cite{qin2022searching}. Moreover, people have also treated large language models as interpreters in code interpretation \cite{firdous2023openai} and generators in essay generation \cite{herbold2023ai}, and found that large language models can handle these tasks effectively.
Motivated by this, in this work we are wondering, \textit{if we can also treat the large language model as an action recognizer in skeleton-based human action recognition?} 
In general, large language models hold some characteristics that are useful for action recognition.
Specifically, pre-trained over a tremendously large corpus \cite{touvron2023llama,brown2020language} that generally contains very rich human-centric behavior descriptions, large language models that hold \textit{large model architectures} could naturally contain rich implicit knowledge \cite{ouyang2022training,qu2024lmc} of human behaviors, and hold strong capability in handling different inputs.
Thus, it can be very promising if we treat the large language model as a human action recognizer.

However, despite the success of large language models in handling various human languages,
it can be challenging to treat them as action recognizers. This is because, while large language models typically take sentences in human languages as input instructions, the input signal of the action recognizer (i.e., the skeleton sequence) is not in a ``sentence format'' that is ``friendly'' to large language models. 
To handle this issue, given a large language model initialized with its pre-trained weights, one potential way is to adjust the weights of the full model to fit the non-linguistic input action signals. Recently, several works \cite{liu2023visual,jiang2023motiongpt} have been proposed to fine-tune large language models and adjust their pre-trained weights to handle non-language tasks. However, as shown in the previous study \cite{weng2023g}, 
the adjustment of the large language model's pre-trained weights can hurt its generalizability and lead to the loss of its pre-learned rich knowledge. This is clearly undesirable here, as we hope the large language model to be knowledgeable in order for it to be an accurate action recognizer.

Taking this into account, in this work, we aim to harness the large language model as an action recognizer, and at the same time keep its pre-trained weights untouched to preserve its pre-learned rich knowledge. 
Specifically, we propose a novel action recognition framework named \textbf{L}arge \textbf{L}anguage \textbf{M}odel as an \textbf{A}ction \textbf{R}ecognizer (\textbf{LLM-AR}).
As shown in Fig.~\ref{fig:intro}, to perform action recognition, our LLM-AR framework first involves a linguistic projection process to project the input action signal into its ``sentence format'', that can be ``friendly'' and compatible to the large language model pre-trained over human sentences.
Then the large language model takes in this ``sentence format'' of the action signal together with a human sentence as the instruction to predict the name of the action.
In the rest of this work, for simplicity, we call the action signal in its ``sentence format'' an ``action sentence''. 

In our framework, to project the action signals into ``action sentences'' (i.e., performing linguistic projection), we incorporate several designs.
Basically, to perform such a process, we observe that, every sentence in a language generally can be regarded as a sequence consisting of discrete word tokens. Inspired by both this observation and previous works \cite{zhang2023motiongpt,zhang2023t2m}, in LLM-AR, we first learn an action-based vector quantized variational autoencoder (VQ-VAE) model to project each action signal into a sequence of discrete tokens.

However, while we can basically represent each ``action sentence'' by a sequence of discrete tokens, it is not necessary that any discrete token sequence can represent an ``action sentence'' well. 
In fact, to enable a large language model to become an accurate action recognizer based on the ``action sentences'', besides being discrete sequences of tokens, ideally, the ``action sentences'' also need to fulfill some requirements as discussed below.
Firstly, since large language models are generally pre-trained over corpus consisting of sentences in human languages, to facilitate large language models in taking ``action sentences'' as instructions together with sentences in human languages, these ``action sentences'' should be ``like'' sentences in human languages so as to be more friendly to the large language model.
On the other hand, ``action sentences'' should still maintain good representations of their original action signals in order for the large language model to accurately perform action recognition based on them.

To better meet the above requirements in the linguistic projection process, we further incorporate our framework with two designs below. Firstly, to make the ``action sentences'' more ``like'' sentences in human languages, we get inspiration from previous linguistic and natural language processing works 
\cite{Zipf:1935kj,piantadosi2014zipf,shieber1985evidence,joshi1990convergence,papadimitriou2023pretrain} which show that, languages as the communication tools of human beings often contain human inductive biases.
Thus, we propose to incorporate our framework with a learning strategy to regularize the projected ``action sentences'' to follow human inductive biases as well. Secondly, to make the projected ``action sentence'' a good representation of its original action signal, motivated by the human skeleton's tree-like nature \cite{wei2017human} and hyperbolic space's superior ability in representing tree structures \cite{ganea2018hyperbolic},  and inspired by previous works \cite{acquavivahyperbolic,li2023resizing}, we further incorporate our action-based VQ-VAE model with a hyperbolic codebook.

Once we finish the learning of the linguistic projection process, we perform low-rank adaptation (LoRA) \cite{hu2021lora} on the large language model to let the model understand the projected ``action sentences''. Note that since the pre-trained weights of the large language model are untouched throughout the LoRA process, we can preserve the pre-learned rich knowledge in the large language model and thus utilize the large language model conveniently in our framework.

The contributions of our work are as follows.
1) 
We propose a novel action recognition framework named LLM-AR. To the best of our knowledge, this work is the first exploration on treating a large language model with its pre-trained weights untouched as an action recognizer. 
2) 
We introduce several designs in our framework to facilitate the linguistic projection process and produce the ``action sentences''.
3) 
LLM-AR achieves state-of-the-art performance on the evaluated benchmarks.

%% file: sec/2_related_work.tex
\section{Related Work}
\label{sec:related}

\noindent\textbf{Human Action Recognition.}  For tackling human action recognition, most of the existing methods can be roughly categorized into two groups: RGB-based methods \cite{wang2016temporal,veeriah2015differential} and skeleton-based methods \cite{yan2018spatial,shi2019skeleton,shi2019two,cheng2020decoupling,chen2021channel}. 
 
As skeleton sequences can represent human behaviors in a succinct yet informative manner, skeleton-based action recognition has received lots of research attention \cite{liu2016spatio,liu2017global,ke2017new,yan2018spatial,shi2019skeleton,shi2019two,cheng2020decoupling,cheng2020skeleton,chen2021channel,cai2023ske2grid,wang2023neural,duan2023skeletr,motionbert2022,foo2023unified,xiang2023generative,xu2023language,wang20233mformer,zhang2021stst,franco2023hyperbolic,chen2022hmanet,leng2023benefit,xin2023skeleton,hachiuma2023unified,xu2022topology,zhang2023hierarchical}. 
In the early days, different works have been proposed to use different network architectures to handle skeleton-based action recognition. Liu et al. \cite{liu2016spatio} proposed to perform skeleton-based action recognition through a spatial-temporal LSTM architecture. Ke et al. \cite{ke2017new} proposed to transform skeleton sequences into grey images and process these images through a CNN network. As time passed, GCN tends to be a popular network architecture \cite{yan2018spatial,li2019actional,shi2019two,cheng2020skeleton,chen2021channel,chi2022infogcn,xiang2023generative,xu2023language}. Yan et al. \cite{yan2018spatial} proposed ST-GCN that made the first attempt in performing skeleton-based action recognition using a GCN architecture. After that, various different GCN-based methods have been further proposed, such as AS-GCN \cite{li2019actional}, 2s-AGCN \cite{shi2019two}, Shift-GCN \cite{cheng2020skeleton}, and CTR-GCN \cite{chen2021channel}. More recently, a number of works \cite{wang20233mformer,foo2023unified,zhang2021stst,motionbert2022} further proposed to perform skeleton-based action recognition via training a model with a large transformer architecture in an end-to-end manner. Some of such methods include 3Mformer \cite{wang20233mformer} and UPS \cite{foo2023unified}.

Different from these approaches, in this work, from a novel perspective, we investigate how to instruct the large language model to perform skeleton-based action recognition while keeping its pre-trained weights untouched to preserve its pre-learned rich knowledge. To achieve this, we involve our framework with a novel linguistic projection process to project each input action signal into an ``action sentence''. Specifically, we incorporate several designs in the linguistic projection process to make the projected ``action sentences'' more ``like'' sentences of human languages so as to make them more friendly to large language models, and meanwhile keep the ``action sentence'' a good representation of the input action signal.

\noindent\textbf{Large Language Models.} 
Recently, a variety of large language models have been proposed, such as GPT \cite{brown2020language} and LLaMA \cite{touvron2023llama}. Pre-trained over a tremendous number of word tokens, these large language models with large model architectures have been shown to contain very rich knowledge \cite{ouyang2022training,qu2024lmc}. Consequently, large language models have been extensively explored in various tasks \cite{jiao2023chatgpt,herbold2023ai,firdous2023openai,zhang2023motiongpt,gong2024signllm,zhu2023llafs,foo2023aigenerated}, such as language translation \cite{jiao2023chatgpt}, essay generation \cite{herbold2023ai}, and code interpretation \cite{firdous2023openai}.
In this work, we design a novel framework treating the large language model as a human action recognizer leveraging our designed linguistic projection process.

%% file: sec/3_method.tex
\section{Proposed Method}
\label{sec:method}

Given a skeleton sequence as the input action signal, the goal of action recognition is to predict its corresponding action class. Recently, holding large model architectures and containing very rich knowledge, pre-trained large language models have been shown to be powerful in handling sentences of human languages, and thus have become useful tools in many natural language processing tasks. Inspired by this, in this work, we aim to \textit{leverage the large language model as an effective action recognizer.} To achieve this, we propose a novel framework \textbf{LLM-AR}. Specifically, LLM-AR first performs a linguistic projection process to project the input action signal (i.e., the skeleton sequence) into an ``action sentence''. After that, LLM-AR passes the ``action sentence'' into the large language model to derive the corresponding human action. Below, we first describe how LLM-AR performs the linguistic projection process, and then introduce the overall training and testing scheme of LLM-AR. 

\subsection{Linguistic Projection}

In our framework, to enable the large language model to perform action recognition, we first perform a linguistic projection process to project each input action signal into an ``action sentence''. To learn to perform such a projection, with the observation that each sentence in human languages is essentially a sequence of discrete word tokens, motivate by previous works \cite{zhang2023motiongpt,zhang2023t2m}, we first learn an action-based VQ-VAE model to project each input action signal into a sequence of discrete tokens. We then further (1) involve the above learning process with a human-inductive-biases-guided learning strategy to make the projected ``action sentences'' more ``like'' sentences in human languages, and (2) incorporate the action-based VQ-VAE model with a hyperbolic codebook to keep ``action sentences'' good representations of the action signals.

\textbf{Action-based VQ-VAE Model.} VQ-VAE models \cite{van2017neural,acquavivahyperbolic,zhang2023regularized}
have been popularly used in converting an image into a discrete token sequence. Inspired by this, in our framework, to convert the input action signal into a sequence of discrete tokens, we first learn an action-based VQ-VAE model.

Specifically, similar to the architecture of previous VQ-VAE models \cite{van2017neural,zhang2023t2m,zhang2023motiongpt}, our action-based VQ-VAE model involves an encoder $E$, a decoder $D$, and a codebook $C$ consisting of $U$ learnable tokens (i.e., $C = \{c_u\}^U_{u=1}$ where $c_u \in \mathbb{R}^{d_u}$). We here keep $d_u$ to be the same as the dimension of the word token in the used large language model, and keep $U$ an even number. Among the aforementioned three components of the model, given an input action signal $s_{1:V}$, where $V$ represents the action length, the encoder $E$ first encodes the action signal through 1D convolutions over the time dimension into a sequence of latent features $f_{1:W}$, where $f_w \in \mathbb{R}^{d_u}$ and $W$ represents the length of the latent feature sequence. After this, to discretize the latent features $f_{1:W}$, a quantization operation is performed to replace each feature $f_w$ with its nearest token in the codebook as:
\begin{equation} \label{eq:quantization}
\setlength{\abovedisplayskip}{3pt}
\setlength{\belowdisplayskip}{3pt}
\begin{aligned}
f^d_w = \underset{c_u \in C}{\mathrm{argmin}} \big(dist(f_w, c_u)\big)
\end{aligned}
\end{equation}
where $f^d_w$ is the discrete version of $f_w$, and $dist(\cdot, \cdot)$ represents a distance function. Finally, the decoder aims to recover the original action signal $s_{1:V}$ from the sequence of discrete latent features (tokens) $f^d_{1:W}$. Through the above process of encoding, quantization, and decoding, we can project an action signal into a sequence of informative discrete tokens $f^d_{1:W}$. The more detailed architecture of this action-based VQ-VAE model is provided in supplementary.

\textbf{Human-inductive-biases-guided Learning Strategy.} 
Above we learn to project each input action signal into a sequence of discrete tokens. 
To enable the learned token sequences to be more friendly to the large language model, here we aim to make these sequences more ``like'' sentences in human languages. To achieve this, we get inspiration from massive existing studies \cite{Zipf:1935kj,piantadosi2014zipf,shieber1985evidence,joshi1990convergence,papadimitriou2023pretrain} which show that, human languages as tools created for human communication naturally contain human inductive biases.  Besides, it is also further shown by the previous study \cite{papadimitriou2023pretrain} that, inputs following human inductive biases are more friendly to large language models. Taking these into consideration, we aim to optimize the set of learned token sequences to also follow human inductive biases, so that these sequences can be more ``like'' human sentences and thus become more friendly to be used by large language models. 

Below, we first introduce the human inductive biases that are generally recognized as being present in human languages. Such biases include: (a) a human language naturally follows the Zipf's law \cite{Zipf:1935kj,piantadosi2014zipf,papadimitriou2023pretrain}, and (b) a human language is generally context-sensitive \cite{shieber1985evidence,joshi1990convergence,papadimitriou2023pretrain}. After introducing these human inductive biases, we then describe our proposed human-inductive-biases-guided learning strategy.

As for the Zipf's law in (a), during daily communications of human beings, there naturally exist some words that are used more commonly and some words more rarely. Zipf's law intuitively represents this imbalanced usage frequency of word tokens in human languages. Specifically, Zipf's law states that, in a human language, the $i$-th most commonly used word has its usage frequency roughly proportional to:
\begin{equation} \label{eq:zipf}
\setlength{\abovedisplayskip}{3pt}
\setlength{\belowdisplayskip}{3pt}
\begin{aligned}
\frac{1}{(i + \beta)^\alpha}
\end{aligned}
\end{equation}
where $\alpha \approx 1$ and $\beta \approx 2.7$ \cite{Zipf:1935kj}. The reason why Zipf's law consistently appears across different human languages is also theoretically analyzed by various works \cite{corominas2011emergence,mandelbrot1954structure} from different perspectives, such as from the evolution of human communications.  

With respect to the context-sensitivity of human languages in (b), intuitively, the context-sensitivity refers to the inductive bias that when people formulate their sentences, they often do not use each word token independently. Instead, their usage of different tokens in formulating a sentence is often correlated. Note that, while human languages are generally believed by linguists to be context-sensitive, how to explicitly represent context-sensitivity remains a difficult problem. 
Here, inspired by \cite{papadimitriou2023pretrain} in its way of representing context-sensitivity, we represent this human inductive bias as follows.
Specifically, given the codebook $C$ consisting of $U$ tokens where $U$ is an even number, during initializing $C$, we first randomly split the codebook into two halves (i.e., $\{1, ...,\frac{U}{2}\}$ and $\{1 + \frac{U}{2}, ..., U \}$). Next, we regard each token $c_u$ in the first half (i.e., $u \in \{1, ..., \frac{U}{2} \}$) and the token $c_{u+\frac{U}{2}}$ in the second half to be a pair of correlated tokens. As shown in \cite{papadimitriou2023pretrain},
such a pairing mechanism is a simple yet very effective way of representing the context-sensitivity bias in human languages.

\begin{figure*}[t]
\centering
\includegraphics[width=\linewidth]{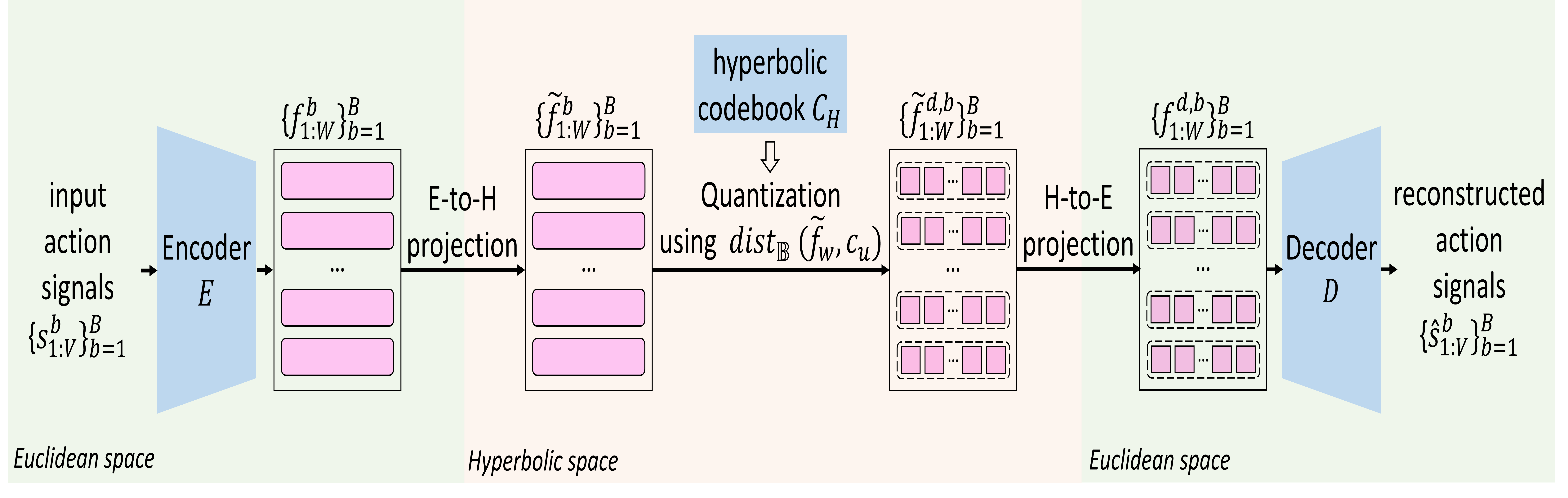}
\vspace{-0.3cm}
\caption{Overview of the action-based VQ-VAE model with the hyperbolic codebook $C_H$ incorporated. Given a batch of input action signals $\{s^b_{1:V}\}_{b=1}^B$, to optimize the action-based VQ-VAE model, $\{s^b_{1:V}\}_{b=1}^B$ are first fed to the encoder $E$ to get the corresponding latent features $\{f^b_{1:W}\}_{b=1}^B$. Next, to leverage the hyperbolic codebook $C_H$ that can serve as a good representation of the tree-like human skeletons to perform quantization, $\{f^b_{1:W}\}_{b=1}^B$ are projected into the hyperbolic space via the process of E-to-H projection. After that, the quantization is performed in the hyperbolic space using $dist_{\mathbb{B}}(\widetilde{f}_w, c_u)$ defined in Eq.~\ref{eq:hyperbolic_distance} as the distance function. Finally, after quantization, the discrete version of the latent features are passed back into the Euclidean space via the process of H-to-E projection to reconstruct the input action signals through the decoder $D$.}
\vspace{-0.35cm}
\label{fig:hyperbolic}
\end{figure*}

To optimize the set of learned token sequences (i.e., ``action sentences'') $\{f^d_{1:W}\}$ to also follow the above biases like human languages, we then aim to (a) regularize the set of ``action sentences'' to follow Zipf's law, as well as (b) regularize each ``action sentence'' to be formulated using more correlated word tokens so that the formulated ``action sentences'' can better follow the context-sensitivity bias.  To achieve this, we design a human-inductive-biases-guided learning strategy that consists of the following steps. (1) Given a batch of action signals $\{s^b_{1:V}\}_{b=1}^B$ with batch size $B$, we first encode these signals through the encoder $E$ of the action-based VQ-VAE model to get their corresponding latent features $\{f^b_{1:W}\}_{b=1}^B$. (2) After that, during discretizing the latent features using tokens $\{c_u\}^U_{u=1}$ in the codebook $C$, we measure the token usage of each sequence of latent features $f^b_{1:W}$ in a differentiable way via the Gumbel Softmax trick \cite{jang2016categorical} as: 
\begin{equation} \label{eq:gumble}
\setlength{\abovedisplayskip}{3pt}
\setlength{\belowdisplayskip}{3pt}
\begin{aligned}
t^b = \sum_{w=1}^W d^b_w, \textbf{where}~d^b_w = \mathrm{Gumble\_Softmax}(-dist(f^b_w, c_u))
\end{aligned}
\end{equation}
where $\mathrm{Gumble\_Softmax}(\cdot)$ is the Gumble Softmax trick, and $d^b_w$ is the one-hot vector with length $U$ as the output of the Gumble Softmax trick. Besides, $t^b$ is a vector with length $U$, and the value of the $u$-th element of $t^b$ represents the number of times token $c_u$ is used in discretizing $f^b_{1:W}$ and formulating its corresponding ``action sentence''.
(3) Next, denoting $D_{Zipf}$ the Zipf distribution with $\alpha = 1$ and $\beta = 2.7$, we regularize the set of ``action sentences'' to follow Zipf's law through $L_{Zipf}$ as:
\begin{equation} \label{eq:zipf_loss}
\setlength{\abovedisplayskip}{3pt}
\setlength{\belowdisplayskip}{3pt}
\begin{aligned}
& L_{Zipf} = JS(D_{freq}\|D_{Zipf}), \textbf{where} D_{freq} = \frac{sort(\sum_{b=1}^B t^b)}{B \times W}
\end{aligned}
\end{equation}
where $sort(\cdot)$ is the sorting operation used to order tokens based on their usage frequency, $D_{freq}$ is the distribution representing the token usage frequency, and $JS(\cdot\|\cdot)$ represents the JS divergence between two distributions. (4) At the same time, we encourage the ``action sentences'' to follow the context-sensitivity bias and use more correlated tokens via $L_{context}$ as:
\begin{equation} \label{eq:context_loss}
\setlength{\abovedisplayskip}{3pt}
\setlength{\belowdisplayskip}{3pt}
\begin{aligned}
& L_{context} = 1 -\frac{\sum_{b=1}^B Corr(t^b)}{B \times W}
\end{aligned}
\end{equation}
where $Corr(t^b)$ leverages min pooling over every pair of elements in $t^b$ (e.g., the pair of $u$-th element and $(u+\frac{U}{2})$-th element) to calculate the number of correlated tokens used in discretizing $f^b_{1:W}$ (more details about this method to measure $Corr(t^b)$ are provided in supplementary).
Note that via minimizing $L_{context}$, we encourage more correlated tokens to be used in formulating ``action sentences''.  
(5) Finally, we formulate the loss function $L_{human}$ that we use to perform our human-inductive-biases-guided strategy as: 
\begin{equation} \label{eq:bias_loss}
\setlength{\abovedisplayskip}{3pt}
\setlength{\belowdisplayskip}{3pt}
\begin{aligned}
& L_{human} = L_{Zipf} + L_{context}
\end{aligned}
\end{equation}
Via incorporating the above learning strategy into the learning process of the action-based VQ-VAE model, 
we can regularize the ``action sentences'' to better follow the human inductive biases that are present in human languages and thus make them more ``like'' sentences in human languages.

Above we inject human inductive biases into ``action sentences''. Here, we notice that, besides inductive biases that can be explicitly described, human languages can also present other implicit characteristics. Thus, to enable ``action sentences'' to be more friendly to large language models which are generally pre-trained over various human languages, we aim to further incorporate implicit characteristics into ``action sentences'' as well. Specifically, since large language models such as LLaMA typically use word tokens $\{c_{LLM}\}$ stored in its first layer to formulate sentences in human languages, we here further align (1) the tokens $\{c_u\}_{u=1}^U$ in codebook $C$ that are used to formulate the ``action sentences'' with (2) the word tokens $\{c_{LLM}\}$ used in the large language model. To achieve such an alignment, we leverage Maximum Mean Discrepancy (MMD) \cite{gretton2006kernel} as an effective feature alignment technique to measure the discrepancy between $\{c_u\}_{u=1}^U$ and $\{c_{LLM}\}$. Specifically, the less $\mathrm{MMD}(\{c_u\}_{u=1}^U, \{c_{LLM}\})$ is, $\{c_u\}_{u=1}^U$ and $\{c_{LLM}\}$ can be regarded as more aligned. We then incorporate this alignment procedure into our proposed strategy via rewriting $L_{human}$ in Eq.~\ref{eq:bias_loss} as:
\begin{equation} \label{eq:total_loss}
\setlength{\abovedisplayskip}{3pt}
\setlength{\belowdisplayskip}{3pt}
\begin{aligned}
& L_{human} = L_{Zipf} + L_{context} + \mathrm{MMD}(\{c_u\}_{u=1}^U, \{c_{LLM}\})
\end{aligned}
\end{equation}
By incorporating this alignment procedure into the previously mentioned learning strategy, we can then lead the formulated ``action sentences'' $\{f^d_{1:W}\}$ to be more friendly to be used by large language models.

\textbf{Hyperbolic Codebook.} Besides making the ``action sentences'' more friendly to the large language model, we also aim to keep them good representations of the original input action signals (i.e., the skeleton sequences). Motivated by the tree-like nature of human skeletons \cite{wei2017human}, we aim to make the word tokens in the ``action sentences'' to well represent such a structure. To achieve this, inspired by the superior capability of the hyperbolic space in embedding tree structures \cite{ganea2018hyperbolic} and motivated by previous VQ-VAE works \cite{acquavivahyperbolic,li2023resizing}, we here further incorporate our action-based VQ-VAE model with a hyperbolic codebook utilizing the Poincar\'e ball model \cite{ganea2018hyperbolic,nickel2017poincare}. The Poincar\'e ball model is an isometric model that can represent the hyperbolic space. Formally, the $n$-dimensional Poincar\'e ball model is defined as $(\mathbb{B}_c^n, g^c_\textbf{x})$, where $\mathbb{B}_c^n = \{\textbf{x} \in \mathbb{R}^n : c\|\textbf{x}\| < 1\}$, $g^c_\textbf{x} = (\gamma^c_\textbf{x})^2I_n$ is the Riemannian metric tensor, $\gamma^c_\textbf{x} = \frac{2}{1-c\|\textbf{x}\|^2}$ is the conformal factor, $I_n$ is the Euclidean metric tensor, and $c$ is the curvature.

Denoting the hyperbolic codebook $C_H = \{c_u\}^U_{u=1}$, where $c_u \in \mathbb{B}^{d_u}_c$, we make the following three changes to incorporate $C_H$ into our action-based VQ-VAE model. We also illustrate where these changes take place in our action-based VQ-VAE model in Fig.~\ref{fig:hyperbolic}. \textbf{(1) Euclidean-to-Hyperbolic (E-to-H) Projection.} Given a latent feature $f_w$ which is originally in the Euclidean space, to convert it into a discrete token using a hyperbolic codebook, we first need to project $f_w$ into the hyperbolic space. Specifically, following \cite{ganea2018hyperbolic}, we perform such a projection through the exponential map $\mathrm{exp}^c_\textbf{0}(\cdot)$ as:
\begin{equation} \label{eq:exp_map}
\setlength{\abovedisplayskip}{3pt}
\setlength{\belowdisplayskip}{3pt}
\begin{aligned}
\widetilde{f}_w = \mathrm{exp}^c_\textbf{0}(f_w) = tanh(\sqrt{c}\|f_w\|)\frac{f_w}{\sqrt{c}\|f_w\|}
\end{aligned}
\end{equation}
where $\widetilde{f}_w$ is the projected feature of $f_w$ in the hyperbolic space. 
\textbf{(2) Hyperbolic Distance Calculation.} Given the projected feature $\widetilde{f}_w$, to perform quantization of $\widetilde{f}_w$ based on the tokens in $C_H$, we need a distance function defined in the hyperbolic space. Here, for simplicity, we use the popularly used \textit{geodesic/induced distance} \cite{ganea2018hyperbolic,khrulkov2020hyperbolic}. Denote this distance
as $dist_{\mathbb{B}}(\cdot, \cdot)$. $dist_{\mathbb{B}}(\widetilde{f}_w, c_u)$ between $\widetilde{f}_w$ and $c_u$ can be defined as:
\begin{equation} \label{eq:hyperbolic_distance}
\setlength{\abovedisplayskip}{3pt}
\setlength{\belowdisplayskip}{3pt}
\begin{aligned}
dist_{\mathbb{B}}(\widetilde{f}_w, c_u) = arccosh \big(1 + 2\frac{\|\widetilde{f}_w - c_u\|^2}{(1 - \|\widetilde{f}_w\|^2)(1 - \|c_u\|^2)}\big)
\end{aligned}
\end{equation}
\textbf{(3) Hyperbolic-to-Euclidean (H-to-E) Projection.} After quantization, we need to project $\widetilde{f}^d_w$ (the discrete version of $\widetilde{f}_w$) back into the Euclidean space in order for it to be passed to the decoder $D$. To achieve this, we follow \cite{ganea2018hyperbolic} to leverage the logarithmic map $\mathrm{log}^c_\textbf{0}(\cdot)$ as:
\begin{equation} \label{eq:log_map}
\setlength{\abovedisplayskip}{3pt}
\setlength{\belowdisplayskip}{3pt}
\begin{aligned}
f^d_w = \mathrm{log}^c_\textbf{0}(\widetilde{f}^d_w) = arctanh(\sqrt{c}\|\widetilde{f}^d_w\|)\frac{\widetilde{f}^d_w}{\sqrt{c}\|\widetilde{f}^d_w\|}
\end{aligned}
\end{equation}
where $f^d_w$ represents the discrete version of the feature projected back in the Euclidean space. With the above changes made, we can seamlessly involve the hyperbolic codebook $C_H$ into our action-based VQ-VAE model, and make the projected ``action sentence'' a better representation of the input action signal.

\subsection{Overall Training and Testing}
\label{sec:overall}

Above we describe how we perform the linguistic projection process in our framework to project each input action signal into an ``action sentence''.
Below, we introduce the overall training and testing scheme of our framework.

\textbf{Training.} The training process of our framework consists of the following two stages: (1) optimizing the action-based VQ-VAE model incorporated with the hyperbolic codebook $C_H$ to acquire the ability to project each action signal $s_{1:V}$ into its corresponding ``action sentence'' $\widetilde{f}^d_{1:W}$; and (2) perform low-rank adaptation (LoRA) \cite{hu2021lora} on the large language model for it to better understand the projected ``action sentences''.

At the first stage, to optimize the action-based VQ-VAE model, given a batch of action signals, we first follow previous VQ-VAE works \cite{van2017neural,zhang2023t2m,zhang2023motiongpt} to use the following loss functions, including the reconstruction loss, the embedding loss, and the commitment loss. Among them, the reconstruction loss $L_{re}$ is designed to regularize the difference between the input action signal and the action signal reconstructed from the decoder $D$, the embedding loss $L_{embed}$ is designed to learn the tokens in the codebook, and the commitment loss $L_{commit}$ is designed to encourage each of the encoded latent features to stay close to its discrete version. Formally speaking, denoting $\{s^b_{1:V}\}_{b=1}^B$ a batch of input action signals, $\{\hat{s}^b_{1:V}\}_{b=1}^B$ the corresponding batch of reconstructed action signals, $\{f^b_{1:W}\}_{b=1}^B$ the latent features encoded from $\{s^b_{1:V}\}_{b=1}^B$, and $\{f^{d,b}_{1:W}\}_{b=1}^B$ the discrete versions of these latent features, the above three loss functions can be written as:
\begin{equation} \label{eq:three_losses}
\setlength{\abovedisplayskip}{3pt}
\setlength{\belowdisplayskip}{3pt}
\begin{aligned}
& L_{re} = L_1^{smooth}(\{s^b_{1:V}\}_{b=1}^B, \{\hat{s}^b_{1:V}\}_{b=1}^B) \\
& L_{embed} = \|sg(\{f^b_{1:W}\}_{b=1}^B) - \{f^{d,b}_{1:W}\}_{b=1}^B\|^2 \\
& L_{commit} = \|\{f^b_{1:W}\}_{b=1}^B - sg(\{f^{d,b}_{1:W}\}_{b=1}^B)\|^2
\end{aligned}
\end{equation}
where $L_1^{smooth}(\cdot)$ represents the L1 smooth loss and $sg(\cdot)$ is the stop-gradient operation that is used to prevent the gradient from flowing through its operand. In addition, to make the learned ``action sentences'' more like sentences in human languages, we also incorporate the above three loss functions with $L_{human}$ defined in Eq.~\ref{eq:total_loss}. Overall, we can write the total loss $L_{total}$ for the first training stage as:
\begin{equation} \label{eq:total_loss_first}
\setlength{\abovedisplayskip}{3pt}
\setlength{\belowdisplayskip}{3pt}
\begin{aligned}
L_{total} =  L_{re} + L_{embed} + \omega_1 L_{commit} + \omega_2 L_{human} 
\end{aligned}
\end{equation}
where $\omega_1$ and $\omega_2$ are weighting hyperparameters. 

We then perform LoRA \cite{hu2021lora} on the large language model in the second stage in order for it to better understand the ``action sentences'' while keeping the pre-trained weights of the model untouched. Specifically, for each training sample consisting of an input action signal and its corresponding ground-truth action, we perform the following steps. 
(1) We first project the action signal into an ``action sentence'' of discrete tokens through the action-based VQ-VAE model learned in the first stage.
(2) We then instruct the large language model to act as an action recognizer via a simple instruction as: ``\texttt{Given a sequence of action tokens [tokens], please predict the corresponding action.}'', where [tokens] represent the word tokens of the ``action sentence'' derived in step (1). 
(3) Finally, during LoRA process, we pass the above instruction into the large language model and encourage the similarity between the tokens $t_p$ predicted by the large language model and the tokens $t_g$ representing the ground-truth action as:
\begin{equation} \label{eq:loss_second}
\setlength{\abovedisplayskip}{3pt}
\setlength{\belowdisplayskip}{3pt}
\begin{aligned}
L_{LoRA} =  L_{ce}(t_p, t_g) 
\end{aligned}
\end{equation}
where $L_{ce}(\cdot, \cdot)$ represents the cross-entropy loss.

\textbf{Testing.} During testing, for each testing action signal, we first use the learned action-based VQ-VAE model to derive its corresponding ``action sentence''. After that, we use the same instruction as in step (2) above to instruct the large language model to predict the corresponding action.

%% file: sec/4_experiments.tex
\section{Experiments}
\label{sec:experiment}

To evaluate the efficacy of our framework, we conduct experiments on 4 datasets including NTU RGB+D, NTU RGB+D 120, Toyota Smarthome, and UAV-Human. 

\subsection{Datasets}

\noindent\textbf{NTU RGB+D} \cite{shahroudy2016ntu} is a large-scale dataset popularly used in human action recognition. It contains around 56k skeleton sequences from 60 activity classes. On this dataset, following \cite{shahroudy2016ntu}, we evaluate our method under the Cross-Subject (X-Sub) and Cross-View (X-View) evaluation protocols.

\noindent\textbf{NTU RGB+D 120} \cite{liu2019ntu} is an extension of the NTU RGB+D dataset. It consists of more than 114k skeleton sequences across 120 activity classes. Following \cite{liu2019ntu}, we evaluate our method on this dataset under the Cross-Subject (X-Sub) and Cross-Setup (X-Set) evaluation protocols.

\noindent\textbf{Toyota Smarthome} \cite{das2019toyota} contains 16,115 video samples over 31 activity classes. On this dataset, we use the skeleton sequences pre-processed by \cite{yang2021selective} and we follow it to evaluate our method under the Cross-Subject (X-Sub) and two Cross-View (X-View1 \& X-View2) evaluation protocols.

\noindent\textbf{UAV-Human} \cite{li2021uav} is a dataset that is captured by unmanned aerial vehicles (UAV). It contains more than 20k skeleton sequences over 155 activity classes, and it is collected from 119 distinct subjects. Following \cite{li2021uav}, we use 89 subjects for training and 30 subjects for testing.

\subsection{Implementation Details}

We conduct our main experiments on Nvidia V100 GPU and we use LLaMA-13B \cite{touvron2023llama} as the large language model. 
During the training process of the action-based VQ-VAE model, we optimize the model for 300,000 iterations using the AdamW \cite{loshchilov2017decoupled} optimizer with an initial learning rate of 2e-4. We set the batch size $B$ to 256, the number of tokens $U$ to 512, and the curvature $c$ of the hyperbolic codebook to 1. Additionally, we set the dimension of each word token $d_u$ to the same size (5120) as the word token in LLaMA-13B. Moreover, we set $w_1$ to 0.02 following previous VQ-VAE works \cite{zhang2023t2m,ng2023can} and set $w_2$ to 0.2. Besides, we set the length ($W$) of each sequence of latent features to be a quarter of the length ($V$) of its corresponding input action signal. During the LoRA process, we develop our code based on the Github repo \cite{lit-lamma}, set the number of iterations to 75,000, and use the AdamW optimizer with an initial learning rate of 3e-3. Moreover, we set the batch size to 256, partitioned into micro-batches of 4. Besides, we set the two hyperparameters of the LoRA process, i.e., $r_{LoRA}$ and $\alpha_{LoRA}$, to 64 and 16 respectively.

\begin{table}[t]
\tiny
\caption{Performance comparison on the NTU RGB+D and NTU RGB+D 120 datasets.}
\vspace{-0.2cm}
\centering
\resizebox{\linewidth}{!}{
\begin{tabular}{ccccc}
\hline
\multirow{2}{*}{Method}
& \multicolumn{2}{c}{NTU RGB+D} 
& \multicolumn{2}{c}{NTU RGB+D 120} \\
\cmidrule(lr){2-3} \cmidrule(lr){4-5} 
& X-Sub & X-View & X-Sub & X-Set \\ \hline
ST-GCN \cite{yan2018spatial} & 85.7 & 92.4 & 82.1 & 84.5 \\
Shift-GCN \cite{cheng2020skeleton}  & 87.8 & 95.1 & 80.9 & 83.2 \\
InfoGCN \cite{chi2022infogcn}  & 89.8 & 95.2 & 85.1 & 86.3 \\
PoseC3D \cite{duan2022revisiting}  & 93.7 & 96.5 & 85.9 & 89.7 \\
FR-Head \cite{zhou2023learning}  & 90.3 & 95.3 & 85.5 & 87.3 \\
Koopman \cite{wang2023neural}  & 90.2 & 95.2 & 85.7 & 87.4 \\
GAP \cite{xiang2023generative} & 90.2 & 95.6 & 85.5 & 87.0 \\
HD-GCN \cite{lee2023hierarchically} & 90.6 & 95.7 & 85.7 & 87.3 \\
STC-Net \cite{lee2023leveraging}  & 91.0 & 96.2 & 86.2 & 88.0 \\
DSTformer \cite{motionbert2022}  & 93.0 & 97.2 & - & - \\
SkeleTR \cite{duan2023skeletr} & 94.8 & 97.7 & 87.8 & 88.3 \\
\hline
Ours & \textbf{95.0} & \textbf{98.4} & \textbf{88.7} & \textbf{91.5} \\
\hline
\end{tabular}}
\label{Tab:nturgb}
\end{table}

\subsection{Comparison with State-of-the-art Methods}

On NTU RGB+D and NTU RGB+D 120 datasets, following the experimental setting of recent works \cite{duan2023skeletr,motionbert2022}, we only use the joint modality of human skeletons during our experiments. We report the results in Tab.~\ref{Tab:nturgb}. As shown, compared to existing skeleton-based action recognition methods, our method consistently achieves the best performance across all the evaluation protocols. This demonstrates the effectiveness of our method. Besides, we also report results on the Toyota Smarthome dataset in Tab.~\ref{Tab:toyota}, and on the UAV-Human dataset in Tab.~\ref{Tab:uav}. As shown, our method consistently achieves the best performance on these datasets. This further shows the efficacy of our method.

\begin{table}[t]
\parbox{0.64\linewidth}{
\caption{Performance comparison on the Toyota Smarthome dataset.}
\vspace{0.2cm}
\centering
\resizebox{\linewidth}{!}{
\begin{tabular}{cccc}
\hline
Method & X-Sub & X-View1 & X-View2 \\ \hline
2S-AGCN \cite{shi2019two} & 58.8 & 32.2 & 57.9 \\
SSTA-PRS \cite{yang2021selective} & 62.1 & 22.8 & 54.0 \\
UNIK \cite{yang2021unik} & 62.1 & 33.4 & 63.6 \\
ML-STGNet \cite{zhu2022multilevel} & 64.6 & 29.9 & 63.5 \\
\hline
Ours & \textbf{67.0} & \textbf{36.1} & \textbf{66.6}\\
\hline
\end{tabular}}
\label{Tab:toyota}
}
\hspace{0.01\linewidth}
\parbox{0.33\linewidth}{
\caption{Performance comparison on the UAV-Human dataset.} 
\vspace{-0.2cm}
\centering
\resizebox{\linewidth}{!}{
\begin{tabular}{cc}
\hline
Method & X-Sub \\ \hline
ST-GCN \cite{yan2017mind} & 30.3 \\
2S-AGCN \cite{shi2019two} & 34.8 \\
Shift-GCN \cite{cheng2020skeleton} & 38.0 \\
ACFL \cite{wang2022skeleton} & 44.2 \\
\hline
Ours & \textbf{46.3}\\
\hline
\end{tabular}}
\label{Tab:uav}
}
\end{table}

\subsection{Ablation Studies}

We conduct extensive ablation experiments on the X-Set protocol of the NTU RGB+D 120 dataset. \textbf{More ablation studies such as experiments w.r.t. hyperparameters are in supplementary.}

\setlength{\columnsep}{0.15in}

\begin{wraptable}[8]{r}{0.45\columnwidth}
\vspace{-0.4cm}
\caption{Evaluation on the human-inductive-biases-guided learning strategy.}
\vspace{-0.2cm}
\resizebox{0.45\columnwidth}{!}
{
\small
\begin{tabular}{l|c}
\hline
Method & Accuracy\\
\hline
w/o biases & 87.6\\
\hline
w/o $L_{Zipf}$ & 89.9\\
w/o $L_{context}$ & 89.8\\
w/o MMD alignment & 90.3\\
\hline
LLM-AR & 91.5 \\
\hline
\end{tabular}}
\label{Tab:ablation_study_1}
\end{wraptable}
\noindent\textbf{Impact of the human-inductive-biases-guided learning strategy.} In our framework, to lead the formulated ``action sentences'' to be more friendly to large language models, we design a human-inductive-biases-guided learning strategy consisting of three components (as shown in Eq.~\ref{eq:total_loss}). To evaluate the efficacy of this strategy, we test four variants. In the first variant (\textbf{w/o biases}), we remove the whole strategy (i.e., all its three components $L_{Zipf}$, $L_{context}$, and MMD alignment) from the learning process. In the second variant (\textbf{w/o $L_{Zipf}$}), we still utilize the strategy but remove its $L_{Zipf}$ component. Moreover, in the third variant (\textbf{w/o $L_{context}$}), we remove the $L_{context}$ component from the strategy, whereas in the fourth variant (\textbf{w/o MMD alignment}), we remove the MMD alignment component.
As shown in Tab.~\ref{Tab:ablation_study_1}, compared to our framework, the performance of the first variant drops significantly. This shows the importance of formulating ``action sentences'' like sentences in human languages. Moreover, our framework also outperforms all the other three variants. This further shows the effectiveness of all the three components of our proposed learning strategy.

\begin{wraptable}[6]{r}{0.42\columnwidth}
\vspace{-0.45cm}
\caption{Evaluation on discretizing latent features into ``action sentences''.}
\vspace{-0.2cm}
\resizebox{0.42\columnwidth}{!}
{
\small
\begin{tabular}{l|c}
\hline
Method & Accuracy\\
\hline
w/o discretization & 83.4\\
with discretization & 91.5 \\
\hline
\end{tabular}}
\label{Tab:ablation_discrete}
\end{wraptable}
\noindent\textbf{Impact of discretizing latent features into ``action sentences''.} 
In our framework, we discretize the encoded latent features to formulate ``action sentences'' consisting of discrete word tokens  (\textbf{with discretization}). To valid this design, we test a variant. In this variant (\textbf{w/o discretization}), instead of discretizing the encoded latent features into ``action sentences'', we directly pass these continuous features to the intermediate layers of the large language model. As shown in Tab.~\ref{Tab:ablation_discrete}, our framework with discretization outperforms this variant. This shows the advantage of discretizing latent features into ``action sentences'', which are more ``like'' human sentences consisting of discrete word tokens,
and thus are more friendly to the large language model pre-trained over human sentences.

\begin{wraptable}[6]{r}{0.32\columnwidth}
\vspace{-0.45cm}
\caption{Evaluation on the hyperbolic codebook $C_H$.}
\vspace{-0.2cm}
\resizebox{0.33\columnwidth}{!}
{
\small
\begin{tabular}{l|c}
\hline
Method & Accuracy\\
\hline
w/o $C_H$ & 89.7\\
with $C_H$ & 91.5 \\
\hline
\end{tabular}}
\label{Tab:ablation_study_2}
\end{wraptable}
\noindent\textbf{Impact of the hyperbolic codebook $C_H$.} 
In our framework, we incorporate our action-based VQ-VAE model with a hyperbolic codebook $C_H$ (\textbf{with $C_H$}). To validate the efficacy of $C_H$, we test a variant (\textbf{w/o $C_H$}) in which the codebook is set up in the Euclidean instead of hyperbolic space. As shown in Tab.~\ref{Tab:ablation_study_2}, our framework involving $C_H$ performs better than this variant. This shows the efficacy of $C_H$ in the hyperbolic space, which can facilitate the ``action sentences'' in representing the tree-like-structured input action signals better.

\begin{wraptable}[5]{r}{0.33\columnwidth}
\vspace{-0.4cm}
\caption{Evaluation on the LoRA process.}
\vspace{-0.25cm}
\resizebox{0.33\columnwidth}{!}
{
\small
\begin{tabular}{l|c}
\hline
Method & Accuracy\\
\hline
All tuning & 79.6\\
\hline
LLM-AR & 91.5 \\
\hline
\end{tabular}}
\label{Tab:ablation_study_3}
\end{wraptable}
\noindent\textbf{Impact of the LoRA process.} In our framework, to make the large language model understand the ``action sentences'' while keeping its pre-trained weights untouched to preserve its rich pre-learned knowledge, we tune the large language model through a LoRA process. To validate this scheme, here we also test a variant (\textbf{all tuning}) on A100 GPU. In this variant, during tuning, after initializing the large language model with its pre-trained weights, all the parameters of the model will undergo gradient updates. As shown in Tab.~\ref{Tab:ablation_study_3}, our framework achieves much better performance than this variant. This shows the superiority of our framework in choosing to perform tuning via LoRA, which enables the large language model's pre-trained weights to be untouched and maintains its pre-learned rich knowledge.

\begin{wraptable}[6]{r}{0.33\columnwidth}
\caption{Evaluation on unseen activity classes.}
\vspace{-0.2cm}
\resizebox{0.33\columnwidth}{!}
{
\small
\begin{tabular}{l|c}
\hline
Method & Accuracy\\
\hline
All tuning & 37.7 \\
\hline
LLM-AR & 62.4 \\
\hline
\end{tabular}}
\label{Tab:unseen}
\end{wraptable}
\noindent\textbf{Evaluation on unseen activity classes.} In the main experiments, following \cite{shahroudy2016ntu,yan2018spatial,lee2023hierarchically}, we evaluate our framework on activity classes that have been seen during training. Here, inspired by that large language models naturally could contain rich knowledge beyond the training activity classes used in our experiments, we are curious, \textit{assuming we have a list of testing activity classes unseen during training, can we also use our framework to perform action recognition on these classes?} To answer this question, we first build a new evaluation protocol for unseen activity classes based on the NTU RGB+D 120 dataset. Under this new protocol, during each time of evaluation, we randomly select 3 classes to form the unseen class list (i.e., the list of testing classes), and use the remaining classes as the classes seen during training (i.e., the training classes). Besides, we instruct the large language model as: ``Given a sequence of action tokens [tokens], please predict the corresponding action from [list].'', where [tokens] represent the word tokens of the ``action sentence'' and [list] represents the unseen class list. We then perform the above evaluation for five times and report the average performance. As shown in Tab.~\ref{Tab:unseen}, even testing on activity classes unseen during training, our framework can still achieve a relatively good performance, while the afore-defined \textbf{all tuning} variant performs much worse. This can be analyzed as, large language models could contain rich pre-learned knowledge w.r.t. the list of unseen classes. Thus, our framework that maintains such rich knowledge can still perform promising recognition on these unseen classes, while the \textbf{all tuning} variant that can lose amount of pre-learned knowledge of the large language model would yield a much worse performance. This further shows the advantage of our framework in maintaining the pre-learned rich knowledge of the large language model.

\setlength{\columnsep}{0.3125in}

%% file: sec/5_conclusion.tex
\section{Conclusion}
\label{sec:conclusion}

In this paper, we have proposed a novel action recognition framework LLM-AR. In LLM-AR, we treat the large language model as an action recognizer, and instruct the large language model to perform action recognition using its contained rich knowledge. Specifically, to lead the input action signals (i.e., the skeleton sequences) to be more friendly to the large language model, we first propose a linguistic projection process to project each action signal into an ``action sentence''. Moreover, we also 
introduce several designs to further facilitate this process. Our framework consistently achieves SOTA performance across different benchmarks.